\documentclass[twoside]{article}

\usepackage[accepted]{aistats2019}
\usepackage[round]{natbib}

\usepackage[utf8]{inputenc} 
\usepackage[T1]{fontenc}    
\usepackage{hyperref}       
\usepackage{url}            
\usepackage{booktabs}       
\usepackage{amsfonts}       
\usepackage{nicefrac}       
\usepackage{microtype}      

\usepackage{amsmath}
\usepackage{amssymb}
\usepackage{amsthm}
\usepackage{graphicx}
\usepackage{caption}
\usepackage{epsfig}
\usepackage{import}
\usepackage{subcaption}
\usepackage{empheq}
\usepackage{wrapfig}
\usepackage{booktabs,tabularx}
\newcommand{\ra}[1]{\renewcommand{\arraystretch}{#1}}

\newtheorem{assumption}{Assumption}

\newcommand\independent{\protect\mathpalette{\protect\independenT}{\perp}}
\def\independenT#1#2{\mathrel{\rlap{$#1#2$}\mkern2mu{#1#2}}}

\newcommand{\dataset}{{\cal D}}

\DeclareMathOperator*{\argmax}{arg\,max}

\begin{document}

%
\runningtitle{Semi-Generative Modelling}

%
\runningauthor{von K{\"ugelgen}, Mey, Loog}

\twocolumn[

\aistatstitle{Semi-Generative Modelling:\\ Covariate-Shift Adaptation with Cause and Effect Features}

\aistatsauthor{Julius von K{\"u}gelgen$^{1,2}$ \And Alexander Mey$^3$ \And  Marco Loog$^{3,4}$}


\aistatsaddress{ $^1$Max Planck Institute for\\ Intelligent Systems, Germany \And $^2$Univ. of Cambridge,\\ United Kingdom \And $^3$Delft Univ. of Technology,\\ The Netherlands \And $^4$Univ. of Copenhagen,\\ Denmark} ]

\begin{abstract}
Current methods for covariate-shift adaptation use unlabelled data to compute importance weights or domain-invariant features, while the final model is trained on labelled data only.
Here, we consider a particular case of covariate shift which allows us also to learn from unlabelled data, that is, combining adaptation with semi-supervised learning.
Using ideas from causality, we argue that this requires learning with both causes, $X_C$, and effects, $X_E$, of a target variable, $Y$, and show how this setting leads to what we call a semi-generative model, $P(Y,X_E|X_C,\theta)$. 
Our approach is robust to domain shifts in the distribution of causal features and leverages unlabelled data by learning a direct map from causes to effects.
Experiments on synthetic data demonstrate significant improvements in classification over purely-supervised and importance-weighting baselines. 
\end{abstract}

\section{INTRODUCTION} \label{sec:introduction}
With advances in algorithms and hardware, the amount of high-quality, labelled training data is becoming the bottleneck for many machine learning tasks.
Methods for making good use of available unlabelled data are thus an active area of research with great potential.
Two established methods addressing this issue are semi-supervised learning and domain adaptation.
Semi-supervised learning aims to improve a model of $P(Y|X)$ through a better estimate of the marginal $P(X)$, obtainable via unlabelled data from the \textit{same} distribution \citep{Chapelle:2010:SL:1841234}.
However, due to different data sources, experimental set-ups, or sampling processes, this i.i.d. assumption is often violated in practice \citep{storkey2009training}.
Domain adaptation, on the other hand, aims to adapt a model trained on a source domain (or distribution)  to a \textit{different, but related} target distribution from which no, or only limited, labelled data is available \citep{pan2010survey, quionero2009dataset}.
This situation arises, for example, when training and test sets are not drawn from the same distribution. 

This paper aims to investigate the possibility of semi-supervised learning in a domain adaptation setting, that is, not only adapting but also actively improving a model given unlabelled data from \textit{different} distributions.
Here, we focus on the most commonly used and well-studied assumption in domain adaptation: the covariate-shift assumption \citep{shimodaira2000improving,sugiyama2012machine}.

With $D=0$ and $D=1$ indicating source and target domains respectively, covariate shift states that the difference in distributions arises exclusively as a consequence of a shift in the marginal distributions, $P(X|D=0) \neq P(X|D=1)$, while the conditional, $P(Y|X)$, remains invariant.
Using the domain variable $D$ this assumption can thus be formulated as $Y\independent D|X$.
Assuming that changes in $P(X)$ are caused externally ($D\xrightarrow{}X$)--as opposed to some internal process like, for example, a sampling bias ($X\xrightarrow{}D$ or $Y\xrightarrow{}D$)--\textit{ this covariate-shift assumption thus implicitly treats all features as causal} ($X\xrightarrow{}Y$) \citep{storkey2009training}, for otherwise the v-structure at X ($D\xrightarrow{}X\xleftarrow{}Y$) would introduce a conditional dependence of $Y$ on the domain $D$ given $X$ \citep{koller2009probabilistic}.

Recent work argued that \textit{semi-supervised learning should not be possible in such a causal learning setting} ($X\rightarrow Y$) as $P(X)$ and $P(Y|X)$ should be independent mechanisms in this case \citep{janzing2010causal, scholkopf2012causal}.
In other words, the conditional distributions of each variable given its causes (i.e., its mechanism) represent ``autonomous modules that do not inform or influence each other'' \citep{PetJanSch17}.
In the causal setting, a better estimate of $P(X)$ obtainable from unlabelled data should thus not help to improve the estimate of the independent mechanism $P(Y|X)$.
\textit{With effect features ($Y\xrightarrow{}X$), on the other hand, semi-supervised learning is, in principle, possible} \citep{janzing2015semi}.

\begin{figure}[tb]
	\centering
	\includegraphics[width = 0.8\columnwidth]{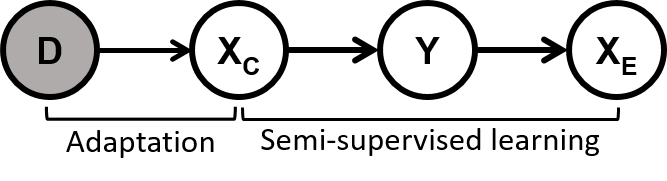}
	\caption{Causal graph of our setting for combining semi-supervised learning and covariate-shift adaptation by learning with both cause- ($X_C$) and effect ($X_E$) features. $D$ indicates the domain, or distribution.}
	\label{fig:overview}
\end{figure}

This need for effect features for semi-supervised learning motivates considering the specific case of covariate shift shown in Fig.~\ref{fig:overview}. 
Note that, by the same v-structure argument as before, we require $D\not \xrightarrow{}X_E$ for covariate shift to hold. 
We thus assume throughout that--through prior causal discovery, expert knowledge, or background information--the underlying causal structure is \textit{known} and compatible with Fig.~\ref{fig:overview}. 
We will make this assumption precise and discuss a possible relaxation in Sec.~\ref{sec:assumptions}.

While requiring particular causal relationships between variables to be known a priori may seem a restrictive assumption, we have already seen that other commonly made, untestable assumptions such as covariate shift also carry implicit assumptions of a causal nature.
Due to the lack of labels from the target distribution, the problem of unsupervised domain adaptation considered in this paper is ill-posed, and thus requires such strong assumptions.
Our assumptions enable us to go beyond adaptation and to explore the possibility of semi-supervised learning away from the i.i.d. setting when the underlying causal structure is known.

The following two examples constitute real-world scenarios which are compatible with the considered setting of prediction from cause and effect features.
\begin{enumerate}
    \item Predicting disease, $Y$, from risk factors like genetic predisposition or smoking, $X_C$, and symptoms, $X_E$:
while we might have (possibly unlabelled) data from multiple geographical regions or demographic groups leading to different distributions over risk factors ($D\rightarrow X_C$), we would not necessarily expect this to affect the behaviour of the disease itself ($X_C\rightarrow Y\rightarrow X_E$).
\item Predicting a hidden intermediate state $Y$ of a physical system with inputs $X_C$ and outputs $X_E$:
 again, we might have data from various experiments with differing input distributions ($D\rightarrow X_C$), but the laws of physics or nature ($X_C\rightarrow Y\rightarrow X_E$) should not change.
\end{enumerate}

We highlight the following contributions:
\begin{itemize}
    \item We introduce the causally-inspired semi-generative model, $P(Y,X_E|X_C,\theta)$, for learning with cause and effect features, and show how its parameters can be fitted from both labelled and unlabelled data in a covariate-shift adaptation setting using a maximum likelihood approach (Sec.~\ref{sec:proposedapproach}).
    \item We empirically demonstrate that our proposed method yields significant reductions in classification error on synthetic data (Sec.~\ref{sec:experiments} \& \ref{sec:discussion}).
    \item We show how our method may also be applied for regression, using real-world protein data (Sec.~\ref{sec:experiments}).
\end{itemize}


\section{RELATED WORK} \label{sec:previouswork}
A sizeable body of literature has been published on the topic of domain adaptation, see e.g. \citep{patel2015visual} for a recent survey. 
Our focus is on {\em unsupervised} domain adaptation under covariate shift where no labels from the target domain are available and the conditional $P(Y|X)$ remains invariant. In general, the aim is to find  a predictor, $f:\mathcal{X}\rightarrow\mathcal{Y}$, which minimizes the target risk, $\mathbb{E}_{P(X,Y|D=1)}L(f(X),Y)$, for a given loss function, $L$. 
Most previous works on this setting fit into one of two families.

Importance weighting approaches make use of the invariance of $P(Y|X)$ to rewrite the unknown target distribution as $P(X,Y|D=1)=w(X)P(X,Y|D=0)$, where the importance weights $w(X)=\scriptstyle\frac{P(X|D=1)}{P(X|D=0)}$ can be estimated from unlabelled data \citep{shimodaira2000improving, sugiyama2007covariate, quionero2009dataset, sugiyama2012machine}.
This allows for empirical risk minimization on the reweighted labelled source sample to approximate the expected target risk.

 Feature transformation approaches, on the other hand, are based on finding domain invariant features in a new (sub)space \citep{SA, GFK}. 
Generally, they learn a map $\phi:\mathcal{X}\rightarrow\mathcal{X}'$ s.t. the projected features are as domain invariant as possible, $P(\phi(X)|D=0)\approx P(\phi(X)|D=1)$. 
Various criteria have been used to measure such similarity, e.g., MMD \citep{TCA}, HSIC \citep{MIDA}, mutual information with $D$ \citep{ITL}, or performance of a domain classifier \citep{DANN}. 
The final model is trained on the transformed labelled sample.

Note that in either approach unlabelled data is used only for adaptation, while the final model is trained on {\em labelled} data only. 
The current work aims to also include {\em unlabelled} data in the model fitting when labelled data is scarce. 
To the best of our knowledge, this is the first work addressing this novel setting.

\section{LEARNING WITH CAUSE AND EFFECT FEATURES} \label{sec:proposedapproach}
We now state our assumptions, show how they lead us to a semi-generative model, and show how to fit its parameters using a maximum-likelihood approach.
Note, however, that our semi-generative model can also be applied in a Bayesian way, see Appendix D of the supplementary material for details and further experiments using a Bayesian approach.

\subsection{Assumptions} \label{sec:assumptions}
Consider the setting of predicting the outcome of target random variable, $Y$, from the observation of two disjoint, non-empty sets of random variables, or features, $X_C$ and $X_E$. 
Assume that we are given a small, labelled sample $\{(x_C^i,y^i,x_E^i)\}_{i=1}^{n_S}$ from a source domain ($D=0$) and a potentially large, unlabelled sample $\{(x_C^j,x_E^j)\}_{j=n_S+1}^{n_S+n_T}$ from a target domain ($D=1$). 
We formalise our causal assumptions as motivated in Sec.~\ref{sec:introduction} using Pearl's framework of a structural causal model (SCM) \citep{pearl2009causality}.

An SCM over a set of random variables $\{X_i\}_{i=1}^d$ with corresponding causal graph $\mathcal{G}$ is defined by a set of structural equations,
\begin{equation*}
X_i := f_i(\mathbf{PA}_{X_i}^{\mathcal{G}}, N_i) \quad \text{for} \quad i=1,\dots,d
\end{equation*}
where $\mathbf{PA}_{X_i}^{\mathcal{G}}$ is the set of causal parents of $X_i$ in $\mathcal{G}$, $N_i$ are mutually independent, random noise variables, and $f_i$ are deterministic functions.
\begin{assumption}[Causal structure] \label{ass:SCM}
The relationship between the random variables $X_C$, $Y$, $X_E$ and the domain indicator $D$ is accurately captured by the SCM
\begin{align}
X_C&:=f_C(D,N_C)\label{eq:scmC}  \\ 
Y&:=f_Y(X_C,N_Y)\label{eq:scmY}\\
X_E&:=f_E(Y,N_E)\label{eq:scmE}
\end{align}
where $N_C$, $N_Y$, and $N_E$ are mutually independent, and $f_C$, $f_Y$, and $f_E$ represent independent mechanisms. 
\end{assumption}
\begin{figure}[t]
    \centering
    \includegraphics[width = 0.8\columnwidth]{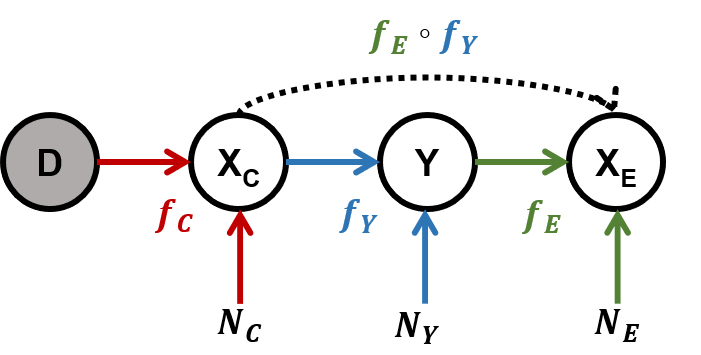}
    \caption{Structural causal model of interest. The dashed arrow illustrates our approach of learning to map $X_C$ to $X_E$ which can be seen as a noisy composition of the mechanisms $f_Y$ and $f_E$.}
    \label{fig:ourSCM}
\end{figure}
This SCM is shown schematically in Fig.~\ref{fig:ourSCM}.
The (unknown) noise distributions together with Eq.~\eqref{eq:scmC}-\eqref{eq:scmE} induce a range of observational and interventional distributions over $(X_C,Y,X_E)$ which depend on $D$. Here, we focus on the two observational distributions arising from the choice of $D$  which we denote by $P(X_C,Y,X_E|D=0)$ (source domain) and $P(X_C,Y,X_E|D=1)$ (target domain).\footnote{Note that even though we focus on the case $D\in \{0,1\}$ here, it should be straight forward to include additional labelled or unlabelled data from different sources as in domain generalisation \citep{rojas2018invariant}.}

It is worth pointing out, that Assumption 1 does not allow a direct causal influence of $X_C$ on $X_E$, and is thus strictly stronger than necessary.
(As stated in Sec.~\ref{sec:introduction}, $D\not \xrightarrow{}X_E$ is sufficient for covariate shift to hold.)
This assumption of two conditionally independent feature sets given $Y$ also plays a key role in the popular co-training algorithm \citep{blum1998combining}.
Interestingly, it has been shown for co-training that performance deteriorates once this assumption is violated and the two feature sets are correlated beyond a certain degree \citep{krogel2004multi}. Similar behaviour can reasonably be expected for our related setting, justifying $X_C\not \xrightarrow{}X_E$.

\subsection{Analysis}
Given that the joint distribution induced by an SCM factorises into independent mechanisms \citep{pearl2009causality},
\begin{equation*}
\textstyle P(X_1,\dots,X_d)=\prod_{i=1}^d P(X_i|\mathbf{PA}_{X_i}^{\mathcal{G}}),   
\end{equation*}
it follows from Assumption 1 that
\begin{equation}\label{eq:facP}
P(X_C,Y,X_E|D)=P(X_C|D)P(Y|X_C)P(X_E|Y).
\end{equation}
It is clear from Eq.~\eqref{eq:facP} that only the distribution of causes is {\em directly} affected by the domain change, while the two mechanisms generating $Y$ from $X_C$, and $X_E$ from $Y$ are invariant across domains. It is this invariance which we will exploit by learning a map from $X_C$ to $X_E$ from unlabelled data, which can be thought of as a noisy composition of $f_Y$ and $f_E$ as indicated by the dashed arrow in Fig.~\ref{fig:ourSCM}.

Note that changes in the distribution of causes are still propagated through the two independent, domain-invariant mechanisms, $P(Y|X_C)$ and $P(X_E|Y)$, and thereby $D$ also {\em indirectly} affects the distributions over $Y$ and $X_E$. 
We also note that for importance weighting it is sufficient to correct for the shift in $X_C$. Writing $w(X_C)=\frac{P(X_C|D=1)}{P(X_C|D=0)}$ it follows from Eq.~\eqref{eq:facP} that
\begin{equation}\label{eq:X_Cweights}
P(X_C,Y,X_E|D=1)=w(X_C)P(X_C,Y,X_E|D=0)
\end{equation}
Thus conditioning on causal features is sufficient to obtain domain-invariance--an idea which also plays a central role in "Causal inference using invariant prediction" \citep{peters2016invariant}.

Since it is the aim of domain adaptation to minimise the target-domain risk, we are interested in obtaining a good estimate of the target conditional, $P(Y|X_C,X_E,D=1)$.
From Eq.~\eqref{eq:facP}, we have
\begin{equation}\label{eq:targetcond}
\begin{aligned}
P(Y|X_C,X_E,D)&=\frac{P(X_C,Y,X_E|D)}{P(X_C,X_E|D)}\\
&=\frac{P(Y|X_C)P(X_E|Y)}{\sum_{y \in \mathcal{Y}}P(y|X_C)P(X_E|y)}.
\end{aligned}
\end{equation}
As the last term does not depend on $D$, this shows that covariate shift indeed holds, as intended by construction. While it would be possible to write the target conditional differently, only conditioning on $X_C$ as in Eq.~\eqref{eq:targetcond} leads to a domain invariant expression. Such invariance is necessary since, due to a lack of target labels, the numerator involving $Y$ can only be estimated in the source domain.

Moreover, Eq.~\eqref{eq:targetcond} shows that the conditional $P(Y|X_C,X_E)$ can be expressed exclusively in terms of the mechanisms $P(Y|X_C)$ and $P(X_E|Y)$, and is thus independent of the distribution over causes, $P(X_C|D)$. 
A better estimate of $P(X_C|D)$ obtainable from unlabelled data will thus not help improve our estimate of $P(Y|X_C,X_E)$.
This is consistent with the claims of \citet{scholkopf2012causal} that the distribution of causal features is useless for semi-supervised learning, while that of effect features may help.
Another way to see this is directly from the data generating process, i.e., the SCM in Assumption 1. 
While Eq.~\eqref{eq:scmC} does not depend on $Y$ (which is only drawn {\em after} $X_C$), Eq.~\eqref{eq:scmE} clearly does.

What is novel about our approach is explicitly considering both cause and effect features at the same time.
Substituting Eq.~\eqref{eq:scmY} into Eq.~\eqref{eq:scmE} we obtain 
\begin{equation*}
X_E=f_E\big(f_Y(X_C,N_Y),N_E\big),
\end{equation*}
so that learning to predict $X_E$ from $X_C$ we may hope to improve our estimates of $f_Y$ and $f_E$.
In terms of the induced distribution, this corresponds to improving our estimates of $P(Y|X_C)$ and $P(X_E|Y)$ via a better estimate of $P(X_E|X_C)$, which we will refer to as the unsupervised model.
This is possible since parameters are shared between the supervised and unsupervised models. 

\subsection{Semi-Generative Modelling Approach}
Our analysis of the different roles played by $X_C$ and $X_E$ suggest explicitly modelling the distribution of $X_E$, while conditioning on $X_C$,
\begin{equation}
P(Y,X_E|X_C,\theta)=P(Y|X_C,\theta_Y)P(X_E|Y,\theta_E),
\end{equation}
where $\theta=(\theta_Y, \theta_E)$.
We refer to the model on the LHS as {\em semi-generative}, as it can be seen as an intermediate between fully generative, $P(X_C,Y,X_E|\theta)$, and fully discriminative, $P(Y|X_C,X_E,\theta)$.

As opposed to a fully-generative model, our semi-generative model is domain invariant due to conditioning on $X_C$ and can thus be fitted using data from both domains.
At the same time, as opposed to a fully-discriminative model, the semi-generative model also allows including unlabelled data by summing (or integrating if $\mathcal{Y}$ is continuous) out $Y$,
\begin{equation}\label{eq:unsupModel}
P(X_E|X_C,\theta)=\sum_{y \in \mathcal{Y}}P(Y=y,X_E|X_C,\theta)
\end{equation}
For our setting, a semi-generative framework thus combines the best from both worlds: domain invariance and the possibility to include unlabelled data in the parameter fitting process.

It is clear from Eq.~\eqref{eq:unsupModel} that we can always obtain the unsupervised model exactly for classification tasks. For regression, however, we are restricted to particular types of mechanisms $P(Y|X_C,\theta_Y)$ and $P(X_E|Y,\theta_E)$ for which the integral can be computed analytically. Otherwise we have to resort to approximating Eq.~\eqref{eq:unsupModel}.

Our approach can then be summarised as follows. 
We train a semi-generative model $P(Y,X_E|X_C,\theta)$, formed by the two mechanisms $P(Y|X_C,\theta_Y)$ and $P(X_E|Y,\theta_E)$, on the labelled sample, such that the corresponding unsupervised model $P(X_E|X_C,\theta)$ (Eq.~\ref{eq:unsupModel}) agrees well with the unlabelled cause-effect pairs.
For prediction, given a parameter estimate $\theta$, the conditional $P(Y|X_C,X_E,\theta)$ can then easily be recovered from $P(Y|X_C,\theta_Y)$ and $P(X_E|Y,\theta_E)$  as in Eq.~\eqref{eq:targetcond}.

\subsection{Fitting by Maximum Likelihood} \label{sec:loglik}
The average log-likelihood of our semi-generative model given the labelled source data is given by
\begin{equation}\label{eq:ell_S}
\ell_S(\theta)
=\frac{1}{n_S}\sum_{i=1}^{n_S} \log P(y^i,x_E^i|x_C^i,\theta)
\end{equation}
and importance-weighting by $w(X_C)$ as described in Eq.~\eqref{eq:X_Cweights} yields the weighted, or adapted, form
\begin{equation}\label{eq:ell_WS}
\ell_{WS}(\theta)
=\frac{1}{n_S}\sum_{i=1}^{n_S} w(x_C^i)\log P(y^i, x_E^i|x_C^i,\theta).
\end{equation}
The corresponding average log-likelihood of the unsupervised model given unlabelled target data is
\begin{equation}\label{eq:ell_T}
    \begin{aligned}
         \ell_T(\theta)
         &=\frac{1}{n_T} \sum_{j=n_S+1}^{n_S+n_T} \log P(x_E^j|x_C^j,\theta)\\
         &=\frac{1}{n_T} \sum_{j=n_S+1}^{n_S+n_T} \log \Big(\sum_{y\in\mathcal{Y}} P(y, x_E^i|x_C^i,\theta) \Big).
    \end{aligned}
\end{equation}
We propose to combine labelled and unlabelled data in a pooled log-likelihood by interpolating between  average source (Eq.~\ref{eq:ell_S}) and target (Eq.~\ref{eq:ell_T}) log-likelihoods,
\begin{equation} \label{eq:ell_P}
\ell_P^{\lambda}(\theta) = \lambda\, \ell_S(\theta) + (1-\lambda)\,\ell_T(\theta),
\end{equation}
where the hyperparameter $\lambda\in [0,1]$ has an interpretation as the weight of the labelled sample.
For example, $\lambda = 1$ corresponds to using only the labelled sample, whereas $\lambda=\frac{n_S}{n_S+n_T}$ gives equal weight to labelled and unlabelled examples, see Sec.~\ref{sec:lambda} for more details.

\section{EXPERIMENTS} \label{sec:experiments}
Since it is our goal to improve model performance with unlabelled data ($n_T$) when the amount of labelled data ($n_S$) is the main limiting factor, we focus in our experiments on the case of small $n_S$ (relative to the dimensionality) and compare learning curves as $n_T$ is increased.

\subsection{Estimators and Compared Methods}
We compare our approach with purely-supervised and importance-weighting approaches which take the known causal structure (Assumption 1) into account:
\begin{itemize}
\item $\hat{\theta}_S=\argmax_{\theta} \ell_S(\theta)$ -- training on the labelled source data only (baseline, no adaptation)
\item $\hat{\theta}_{WS}=\argmax_{\theta} \ell_{WS}(\theta)$ -- training on reweighted source data (adaptation by importance-weighting using known weights on the synthetic datasets)
\item $\hat{\theta}_P^{\lambda}=\argmax_{\theta} \ell_P^{\lambda}(\theta)$ -- training on the entire pooled data set combining unweighted labelled and unlabelled data via $\lambda$ \textbf{(our proposed estimator)}
\end{itemize}
Where applicable, we report the performance of a linear/logistic regression model, $\hat{\theta}_{LR}$, trained on the joint feature set $(X_C,X_E)$, i.e., ignoring the known causal structure. 
Moreover, we also consider $\hat{\theta}_{LR}$ trained after applying different feature transformation methods: TCA \citep{TCA}, MIDA \citep{MIDA}, SA \citep{SA}, and GFK \citep{GFK}. For this we use the domain-adaptation toolbox by Ke Yan with default parameters \citep{DAToolbox}.

\subsection{Synthetic Classification Data} \label{sec:class}
\begin{figure}[]
    \centering
    \includegraphics[width=0.35\textwidth]{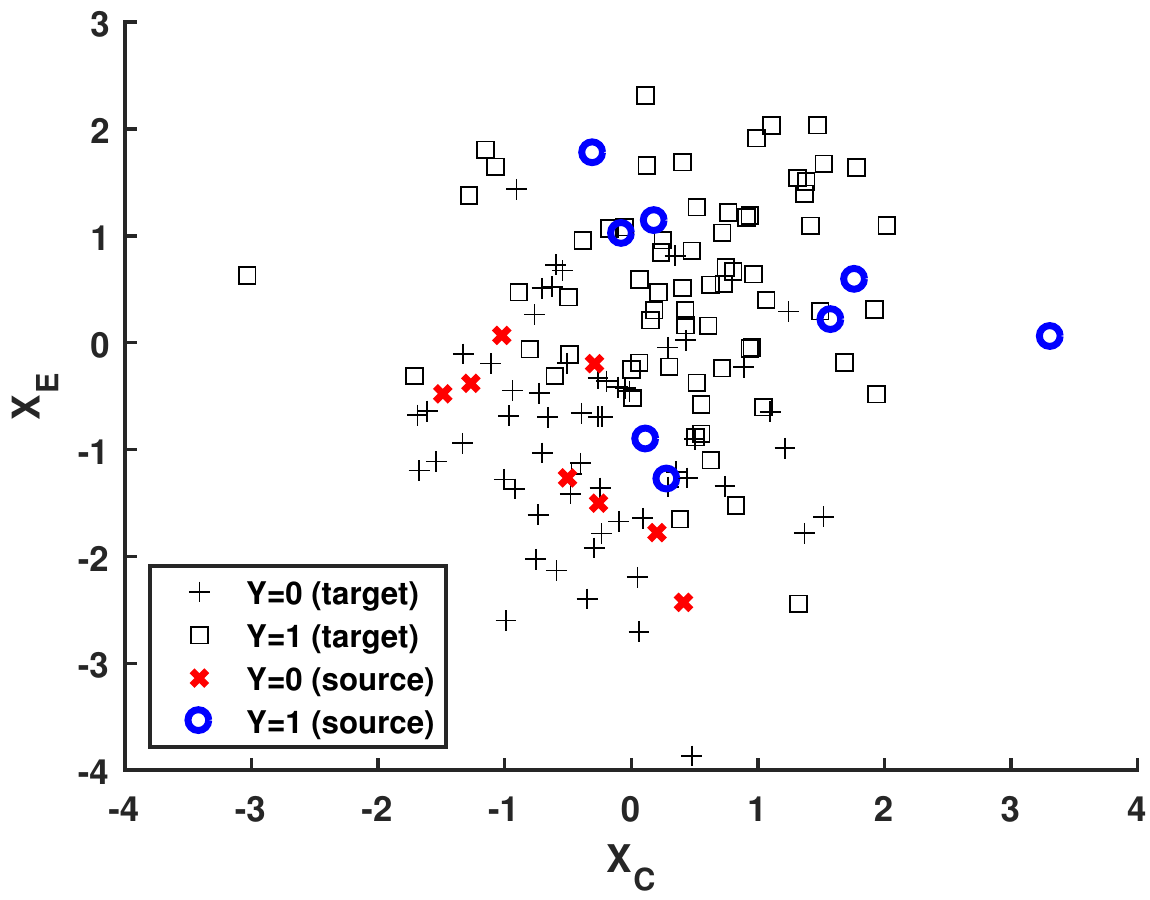}
    \caption{An example of synthetic classification data.}
    \label{fig:classdata}
\end{figure}
To generate synthetic domain-adaptation datasets for binary classification which satisfy the assumed causal structure we draw from the following SCM:
\begin{equation*}\label{eq:classSCM}
    \begin{aligned}
        X_C&:=	\begin{cases} 
        			 \mu_C+\epsilon_C \quad \text{if} \quad D=0,\\
        			 -\mu_C+\epsilon_C  \quad \text{if} \quad D=1,
        		\end{cases}
        		&\epsilon_C \sim \mathcal{N}(0,1)\\[0.5em]
        Y&:=	\begin{cases} 
        			 1 \quad \text{if} \quad \epsilon_Y\leq \sigma(X_C-m),\\
        			 0 \quad \text{if} \quad \epsilon_Y> \sigma(X_C-m),
        		\end{cases}
        		&\epsilon_Y \sim U(0,1)\\[0.5em]
        X_E&:=	\begin{cases} 
        			 \mu_0+\epsilon_E \quad \text{if} \quad Y=0,\\
        			 \mu_1+\epsilon_E  \quad \text{if} \quad Y=1,
        		\end{cases}
        		&\epsilon_E \sim \mathcal{N}(0,1)\\
    \end{aligned}
\end{equation*}
where $\sigma(x)=(1+e^{-x})^{-1}$ is the logistic sigmoid function.
The resulting datasets all have linear decision boundaries, but can differ in domain-discrepancy, class-imbalance, and class-overlap or difficulty, depending on the choice of $\mu_C, m$ and $\mu_{0/1}$, respectively. For one such choice, an example draw is shown in Fig.~\ref{fig:classdata}.

This data generating process induces the  distributions
\begin{equation*}
\begin{aligned}
Y|(X_C=x_C) &\sim \text{Bernoulli}\big(\sigma(x_C-m)\big)\\
X_E|(Y=y) &\sim \mathcal{N}(\mu_y,1).
\end{aligned}
\end{equation*}
The corresponding unsupervised model (Eq.~\ref{eq:unsupModel}) for an unlabelled cause-effect pair $(x_C,x_E)$ is thus given by
\begin{equation}\label{eq:synthunsupmodel}
P(x_E|x_C,\theta)
= \frac{\phi(x_E|\mu_0,1)e^{-(x_C-m)}+\phi(x_E|\mu_1,1)}{1+e^{-(x_C-m)}}
\end{equation}
where $\phi(x|\mu,\sigma^2)$ denotes the pdf of a normal random variable with mean $\mu$ and standard deviation $\sigma$. 
Together with $P(Y|X_C,\theta_Y)$ and $P(X_E|Y,\theta_E)$ given above, Eq.~\eqref{eq:synthunsupmodel} suffices to compute our estimator. Note that, like a logistic regression model, our model has three parameters: $\theta=(m, \mu_0, \mu_1)$.

In addition, to test our approach in a discrete and higher-dimensional setting, we apply our approach to the LUCAS toy dataset\footnote{\url{http://www.causality.inf.ethz.ch/data/LUCAS.html}}, treating 'Lung Cancer' as target $Y$, 'Smoking' and 'Genetics' as causes $X_C$, 'Caughing' and 'Fatigue' as effects $X_E$, and 'Anxiety' as domain indicator $D$.


 
\subsection{Real-World Regression Data}
\begin{figure}[]
	\begin{subfigure}{\columnwidth}
		\centering
		\includegraphics[width=\textwidth, trim={6em 0 6em 0},clip]{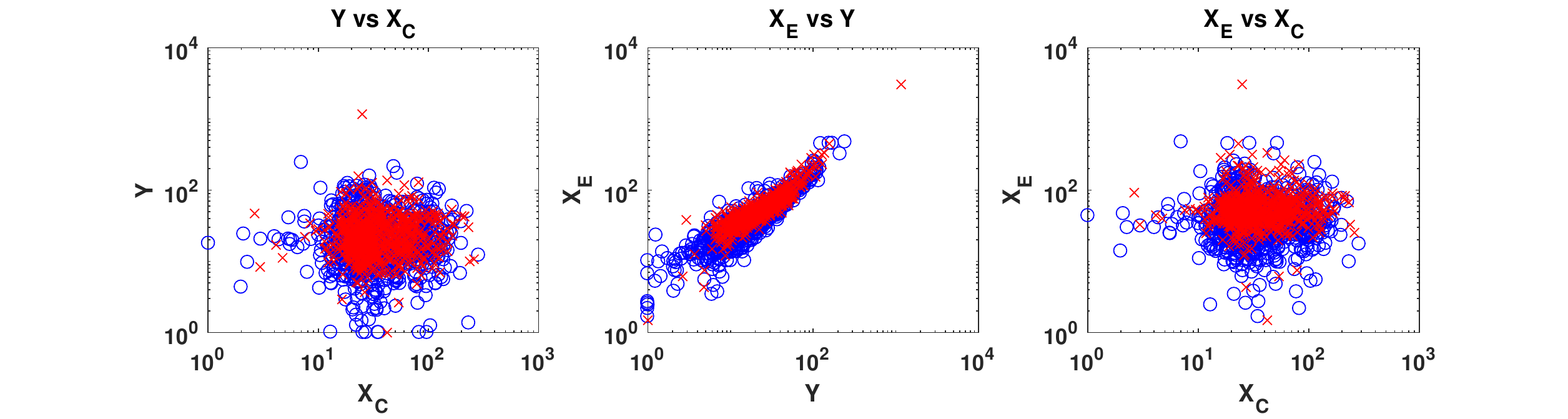}
	\end{subfigure}
    
    \vspace{0.5em}
    \begin{subfigure}{\columnwidth}
		\includegraphics[width=\textwidth, trim={6em 0 6em 0},clip]{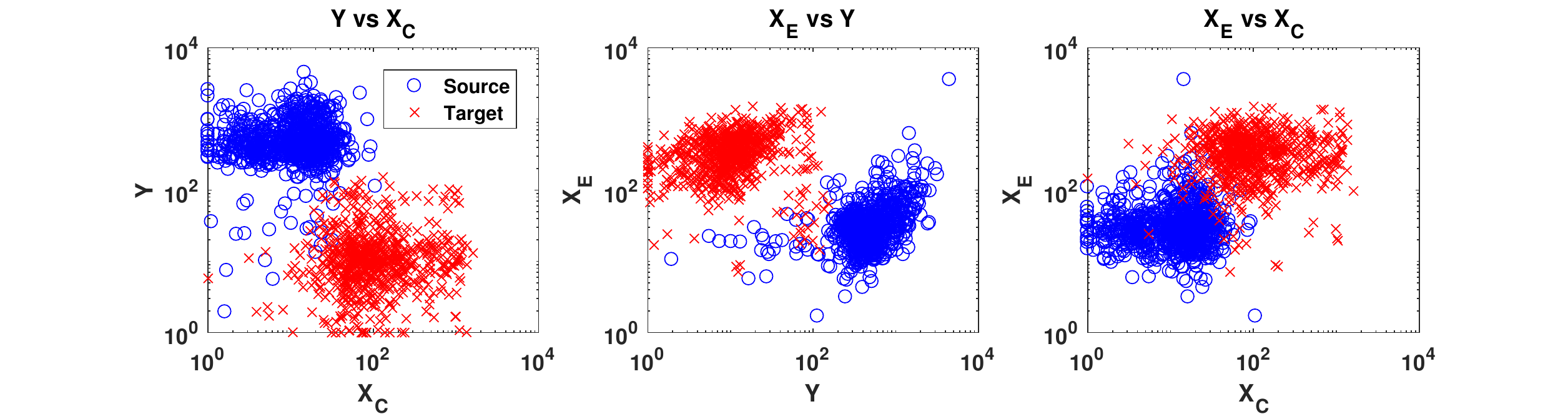}
	\end{subfigure}
    
    \vspace{0.5em}
	\begin{subfigure}{\columnwidth}
	    \centering
	    \includegraphics[scale = 0.275]{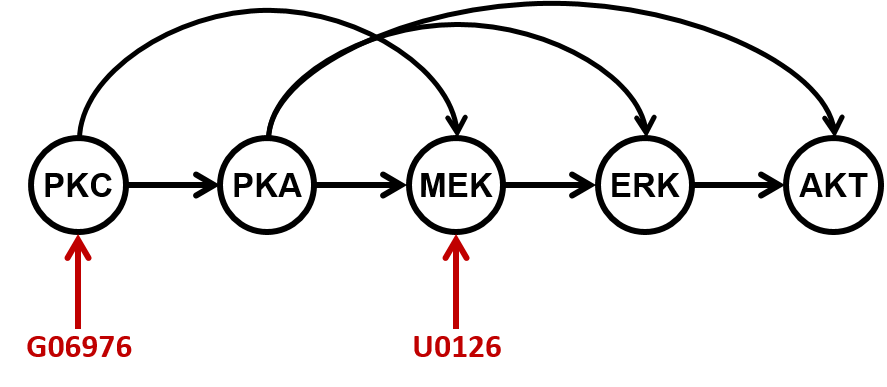}
	\end{subfigure}
	\caption{Protein count data sets for MEK$\xrightarrow{}$ERK$\xrightarrow{}$AKT ($\dataset_1$, top) and PKC$\xrightarrow{}$ PKA$\xrightarrow{}$AKT ($\dataset_2$, middle) in log-log scale. Target domain data is obtained by interventions, shown by red arrows in the inferred causal graph (bottom).}
    \label{fig:realdata}
\end{figure}

To demonstrate how a semi-generative model can be used for linear regression, we apply our approach to the ``Causal Protein-Signaling Network'' data by \citet{sachs2005causal}, which contains single-cell measurements of 11 phospho-proteins and phospho-lipids under 14 different experimental conditions, as well as--important for our method--the corresponding inferred causal graph.
We focus on a subset of variables which seems most compatible with our assumptions\footnote{Assumption 1 is not fully satisfied because of the existence of confounding variables (e.g., PKA, see Fig.~\ref{fig:realdata}), so that conclusions drawn may be limited.
With causal inference and causal structures becoming of interest in more and more areas, however, more suitable real-world data will eventually become abundant.
At this point our work should thus be considered more methodological in nature.
},
and from which we extract two domain adaptation datasets by taking source data to correspond to normal conditions while target data is obtained by intervention on the causal feature, see Fig.~\ref{fig:realdata}. As can be seen, $\dataset_1$ (MEK$\xrightarrow{}$ERK$\xrightarrow{}$AKT) shows a high similarity between domains, whereas $\dataset_2$ (PKC$\xrightarrow{}$ PKA$\xrightarrow{}$AKT) seems more challenging due to high domain discrepancy.

As is often the case with biological data, variables span multiple orders of magnitude and seem to be reasonably-well approximated by power laws.
We therefore decide to first transform the data by taking logarithms and then fit a linear model in log-space, corresponding to a power-law relationship in original space. 
Denoting the log-transformed cause, target, and effect by $X_C,Y$ and $X_E$ as before, and using Gaussian noise with unknown variance, this corresponds to the following model
\begin{equation}\label{eq:regrSCM}
\begin{aligned}
Y&:=a+bX_C+\epsilon_Y, 
&\epsilon_Y \sim \mathcal{N}(0,\sigma_Y^2)\\
X_E&:=c+dY+\epsilon_E,  
&\epsilon_E \sim \mathcal{N}(0,\sigma_E^2), \\
\end{aligned}
\end{equation}
with corresponding distributions
\begin{equation}\label{eq:regrdist}
\begin{aligned}
Y|(X_C=x_C) &\sim \mathcal{N}(a+bx_C,\sigma_Y^2)\\
X_E|(Y=y) &\sim \mathcal{N}(c+dy,\sigma_E^2)
\end{aligned}
\end{equation}

Substituting for $Y$ in the second line of Eq.~\eqref{eq:regrSCM}, and given that the sum of two Gaussian random variables is again Gaussian, we can compute the unsupervised model (Eq.~\ref{eq:unsupModel}) in this case as follows:
\begin{equation} \label{eq:unsupModelRegr}
X_E|(X_C=x_C) \sim \mathcal{N}(c+ad+bdx_C, d^2\sigma_Y^2+\sigma_E^2)
\end{equation}
Eq.~\eqref{eq:regrSCM} and \eqref{eq:unsupModelRegr} combined allow to compute our proposed estimator.
To make predictions given a parameter estimate, we need to compute the $\argmax$ of the conditional (Eq.~\ref{eq:targetcond}).
It is given by
\begin{equation} \label{eq:regrprediction}
\begin{aligned}
\hat{y}
&=\argmax_y P(Y=y|X_C=x_C, X_E=x_E,\theta)\\
&=\frac{\sigma_E^2(a+bx_C)+d^2\sigma_Y^2(\frac{x_E-c}{d})}{\sigma_E^2+d^2\sigma_Y^2}
\end{aligned}
\end{equation}
which can be interpreted as a weighted average of the predictions of each of the two independent mechanisms.
A detailed derivation of Eq.~\eqref{eq:regrprediction} can be found in the supplementary material, Appendix A. 

To investigate how background knowledge can aid our approach in challenging real-world applications, we also fit a model under the constraint  $b,d\leq 0$, that is, fitting lines with negative slope on the harder data set $\dataset_2$. This constraint captures that both PKC$\rightarrow$PKA and PKA$\rightarrow$AKT appear to be inverse relationships--something which may be known in advance from domain expertise. 

\subsection{Choosing the Hyperparameter $\lambda$} \label{sec:lambda}
To choose $\lambda \in [0,1]$, we performed extensive empirical evaluation on synthetic data considering different combinations of $n_S$ and $n_T$, the results of which can be found in the supplement, Appendix C.
For classification, data was generated as detailed in Sec.~\ref{sec:class} with a fixed choice of parameters. For regression, we used a linear Gaussian model to generate synthetic data.

For classification, we found that $\lambda(n_S,n_T)=\frac{n_S}{n_S+n_T}$, giving equal weight to all observations (c.f. Eq.~\ref{eq:ell_P}), i.e., more weight to the unsupervised model as $n_T$ is increased, seems to be a good choice across settings.

In contrast, for linear regression a good choice of $\lambda$ does not seem to depend strongly on $n_S$ and $n_T$.
Rather than weighting all observations equally, values of $\lambda$ giving the fixed majority weight to the average supervised model appear to be preferred.
We thus choose a constant $\lambda = 0.8$ for our regression experiments.
Note, however, that this value can be further increased when more labelled data becomes available (e.g., $\lambda(n_S)=1-\frac{1}{n_S}$) and the unsupervised model becomes obsolete.

\subsection{Simulations and Evaluation}
For synthetic classification experiments, we fix $\mu_C=-1, m=0$ and vary $\mu_0$, and $\mu_1$ as indicated in the figure captions. We thus consider different amounts of labelled data and class-overlap, or difficulty. We perform $10^4$ simulations, each time drawing a new training set of size $(n_S+n_T)$ and a new target-domain test set of size $10^3$. We report test-set averages of error rate and semi-generative negative log-likelihood (NLL), $-\log P(Y,X_E|X_C, \theta)$. The latter is the quantity our model is trained to minimise, and thus acts as a proxy or surrogate for the non-convex, discontinuous 0-1 loss.

For real-world regression experiments, we draw $n_S$ labelled source training data, and reserve 200 target observations as test set. 
From the remaining target data, we then draw $n_T=2,4,...,512$ additional unlablelled training data. 
(Each experiment performed by \citet{sachs2005causal} contains ca. 1000 measurements.)
We perform $10^3$ simulations and report test set averages of root mean squared error (RMSE).

Code to reproduce all our results is available online.\footnote{\url{https://github.com/Juliusvk/Semi-Generative-Modelling}}


\section{DISCUSSION} \label{sec:discussion}
\begin{figure}[]
    \begin{subfigure}{0.45\textwidth}
    	\centering        
        \includegraphics[width=\textwidth]{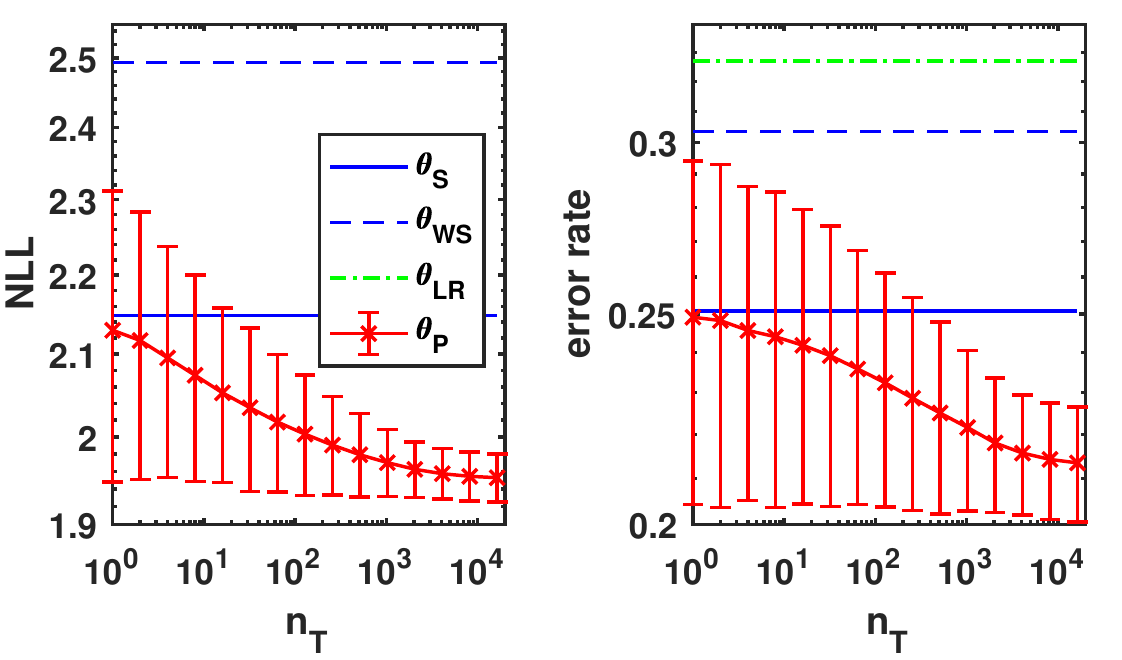} 
        \caption{$n_S=8$, $\mu_1=-\mu_0=0.5$}
        \label{fig:mu05}
    \end{subfigure}
    
    \vspace{0.5em}
    \begin{subfigure}{0.45\textwidth}
    	\centering        
        \includegraphics[width=\textwidth]{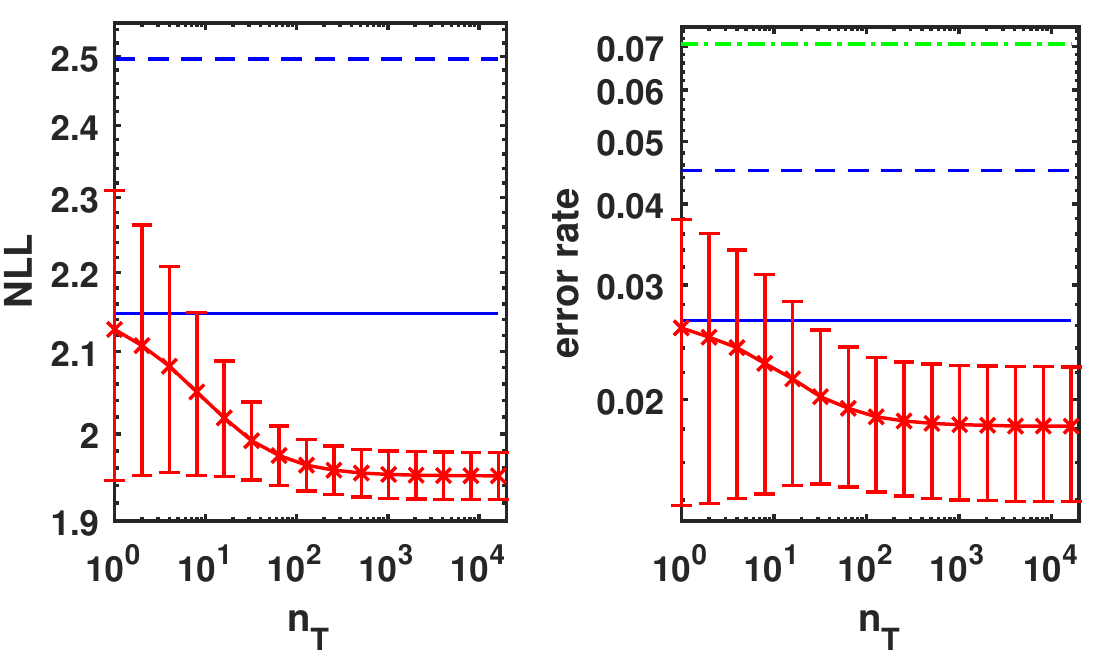}
        \caption{$n_S=8$, $\mu_1=-\mu_0=2$}
        \label{fig:mu2}
    \end{subfigure}
    \caption{Test set averages of negative log-likelihood (NLL) and error rate on synthetic classification data  in log-log scale, using $\lambda=\frac{n_S}{n_S+n_T}$. Error bars indicate one standard deviation. Different values of $\mu_0$ and $\mu_1$ lead to larger (a) or smaller (b) class overlap. This is reflected in the overall error rates. Note that the Bayes error rate in (a) is $\approx0.21$.}
    \label{fig:ClassML}
\end{figure}

Classification results for two synthetic datasets are shown in Fig.~\ref{fig:ClassML}.
For both the more difficult (\ref{fig:mu05}, Bayes error rate $\approx0.21$), and the simpler (\ref{fig:mu2}) data sets, average error rate and variance are monotonically decreasing as a function of $n_T$, leading to significant (paired t-test with $p\ll 0.05$) improvements of $\theta_P$ over $\theta_S$, $\theta_{WS}$, and $\theta_{LR}$ when sufficient unlabelled data is available. 
A very similar behaviour is observed for the semi-generative NLL, indicating that it is a suitable surrogate loss. 
Whereas the largest absolute drop in error rate ($\sim4\%$) is achieved on the more difficult dataset, the largest relative improvement ($\sim30\%$) and earlier saturation occur when--due to the larger absolute value of $\mu_{0/1}$--$X_E$ carries more information about $Y$.
The latter is intuitive as $X_E$ can be interpreted as a second label in this case.

Results for the LUCAS toy data in Table \ref{tab:lucas} show similar behaviour to those in Figure \ref{fig:ClassML}, and demonstrate that our approach is suitable also for discrete data and higher dimensional features.

\begin{table}[]
    \centering
    \ra{1.2}
    \footnotesize
    \caption{Classification test set error rates on the toy LUCAS dataset for $\lambda = n_S/(n_S+\sqrt{n_T})$.}
    \label{tab:lucas}
    \begin{tabularx}{\columnwidth}{@{}lllllll@{}}
        \toprule
        $n_S$\textbackslash$n_T$ & 0 & 1 & 4 & 16 &  64 & 256\\
        \midrule
        8  & 0.232 & 0.230 & 0.226 & 0.220 & 0.212 & 0.208\\
        16 & 0.206 & 0.205  & 0.203  & 0.198 & 0.192 & 0.188\\
       \bottomrule
    \end{tabularx}
\end{table}

\begin{figure}[tb]
    \begin{subfigure}{0.45\columnwidth}
		\centering        
        \includegraphics[width=\textwidth]{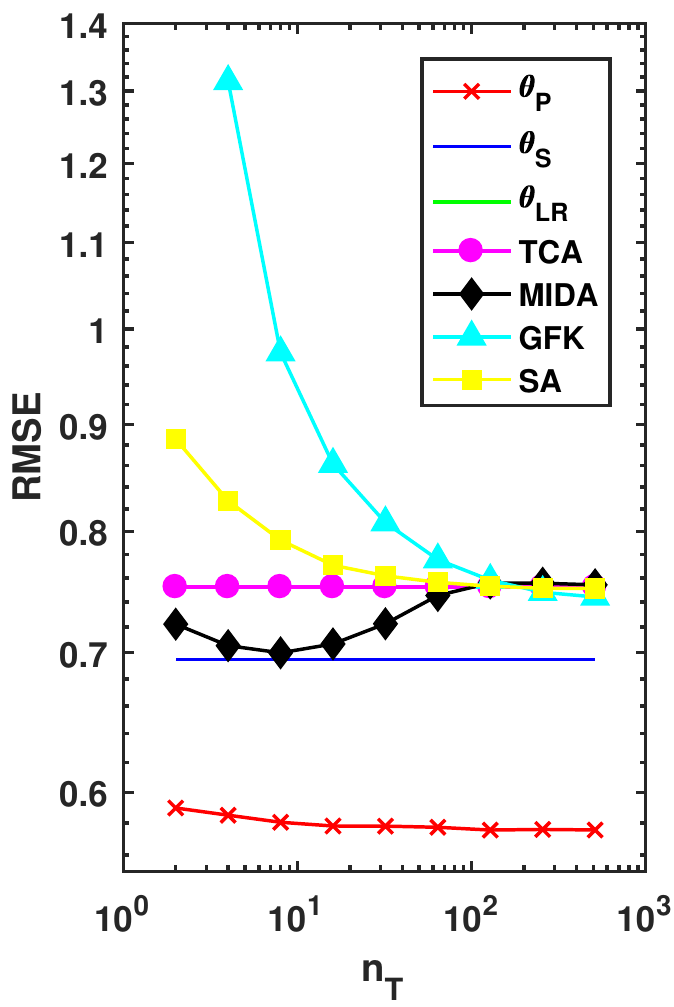} 
        \caption{$\dataset_1$: $n_S=4$}
        \label{fig:resreald1ns4}
    \end{subfigure}
    \hfill
    \begin{subfigure}{0.45\columnwidth}
		\centering        
        \includegraphics[width=\textwidth]{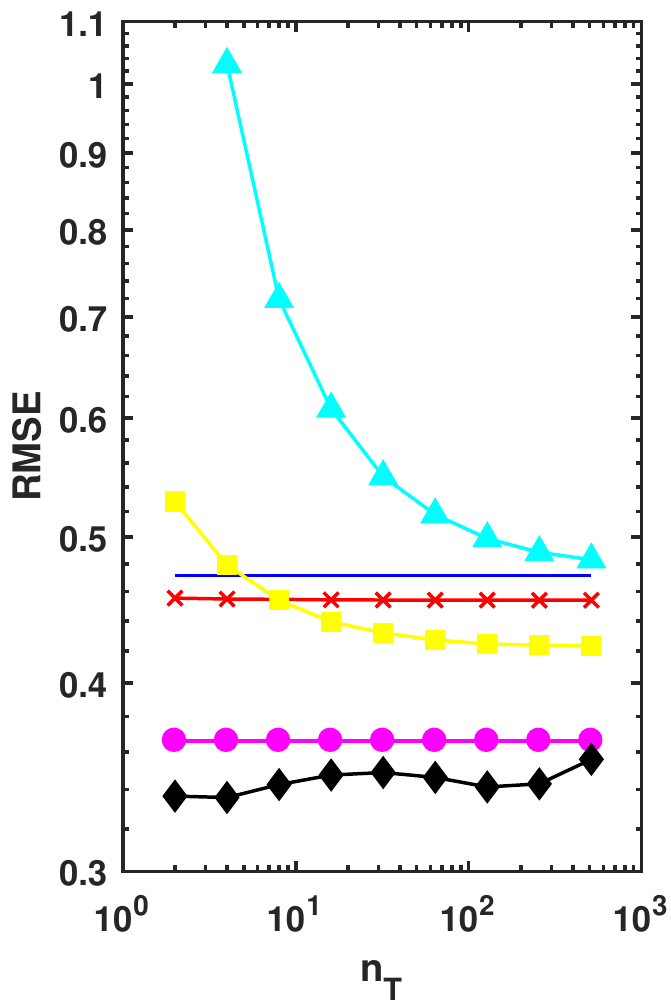} 
        \caption{$\dataset_1$: $n_S=16$}
        \label{fig:resreald1ns16}
    \end{subfigure}

    \vspace{0.5em}
    \begin{subfigure}{0.45\columnwidth}
		\centering        
        \includegraphics[width=\textwidth]{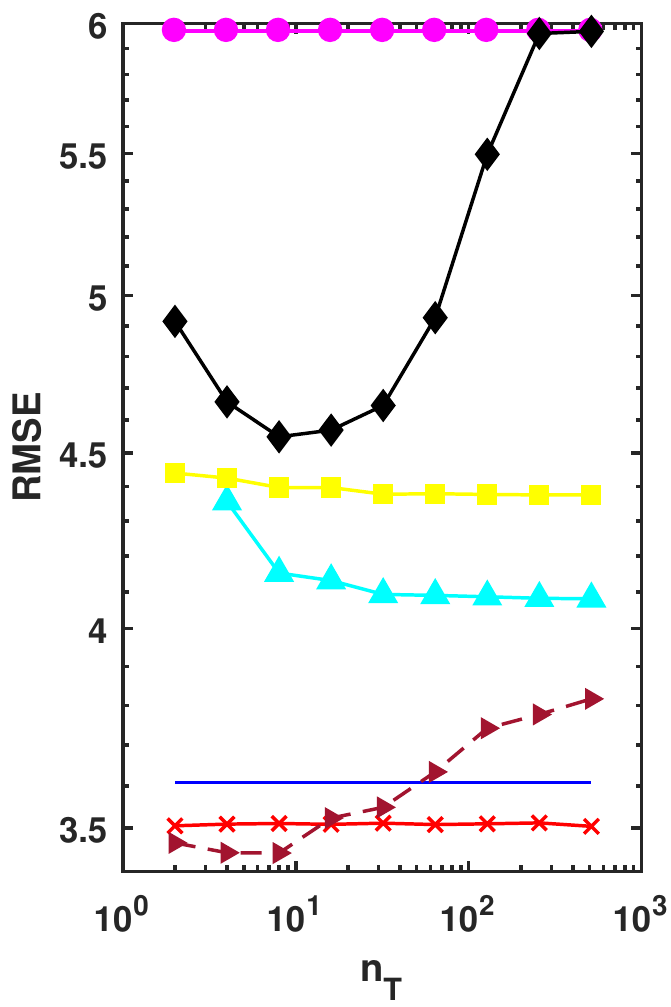}
        \caption{$\dataset_2$: $n_S=4$}
        \label{fig:resreald2ns4}
    \end{subfigure}
    \hfill
    \begin{subfigure}{0.45\columnwidth}
		\centering        
        \includegraphics[width=\textwidth]{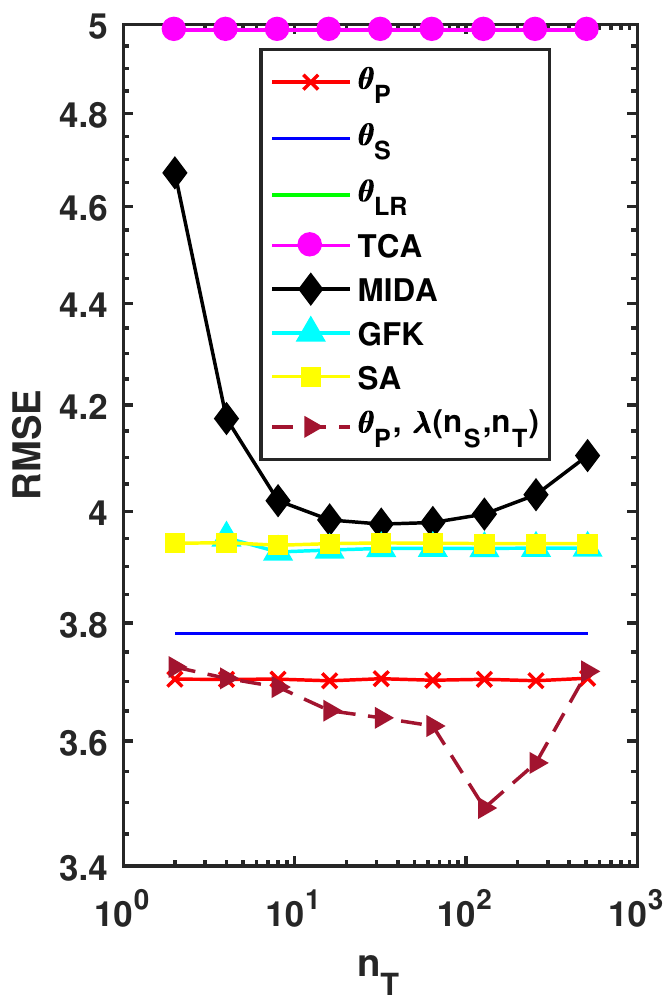} 
        \caption{$\dataset_2$: $n_S=16$}
        \label{fig:resreald2ns16}
    \end{subfigure}
    \caption{Test set averages of RMSE on the real-world regression data sets \citep{sachs2005causal} in log-log scale, using $\lambda=0.8$ except for the dark red curves on $\dataset_2$ which correspond to $\lambda=\frac{n_S}{n_S+n_T}$. On the more difficult dataset $\dataset_2$ (see the higher RMSE), we restricted $\theta_S$ and $\theta_P$ to lines with negative slope.}
    \label{fig:resreal}
\end{figure}

Regression results on the real datasets are shown in Fig.~\ref{fig:resreal}. 
On the simpler $\dataset_1$, our approach outperforms the others when only four labelled observations are available (\ref{fig:resreald1ns4}).
As $n_S$ is increased to 16 (\ref{fig:resreald1ns16}), however, feature transformation methods gain the upper hand.
Given that even $\theta_{LR}$ (coinciding with the curve of TCA) yields better results in this case, a possible explanation is that--due to the common confounder PKA (see Fig.~\ref{fig:realdata})--our assumptions are violated.
On the much more challenging $\dataset_2$, none of the methods yields low RMSE, but the restricted version of our approach performs best, followed by the restricted version of the purely-supervised baseline. 

\paragraph{Comparison with Feature-Transformation Methods}
The case of $\dataset_2$ illustrates a potential advantage of our approach for real-world applications.
Since we use raw features, it is possible to incorporate available domain expertise in the model. Since variables resulting from a transformation of the joint feature set are no longer easily interpretable, including background knowledge is much harder for transformed features.
As such transformations can also introduce new dependencies between variables, it is not clear how our approach and feature transformations can be easily combined.
An interesting idea though could be to relax the assumption $D\not\xrightarrow{}X_E$, and then try to correct for the shift in $X_E$ due to $D$
by learning a transformation of $X_E$ only which maximises 
domain invariance of $\phi(X_E)|X_C$ prior to applying our approach.
As a final note, runtime of our method is roughly an order of magnitude less than for feature-transformation methods.

\paragraph{Combination with Importance Weighting} Importance weighting, on the other hand, should not be seen as an alternative, but rather as complementary to our approach. Through the unlabelled target sample we obtain an estimate of $P(X_C,X_E|D=1)=P(X_C|D=1)P(X_E|X_C)$. The first factor can be used to estimate importance weights, whereas our work has focused on improving the model via information carried by the second factor. Both ideas could be combined by forming a weighted pooled log-likelihood, $\ell_{WP}^{\lambda}$, by replacing $\ell_S$ by $\ell_{WS}$ in Eq.~\eqref{eq:ell_P}. 

\paragraph{Model Flexibility and Role of $\lambda$}
It seems our approach is more promising for classification than for regression tasks. Too much emphasis on the unlabeled data (as controlled by $\lambda$) can, for regression in particular, lead to overfitting of the unsupervised model. This can be observed on $\dataset_2$ for large enough $n_T$ using $\lambda(n_S,n_T)$, and is further illustrated on synthetic data in the supplement, Appendix B.
Since the main difference between regression and classification in our approach is summing over a finite-, or integrating over an infinite number of $y$ when computing the unsupervised model (Eq.~\ref{eq:unsupModel}), we conjecture that model flexibility plays an important role in determining the success of our approach.
If there is a bottleneck at $Y$, so that only few values $y$ can explain a given cause-effect pair $(x_C,x_E)$, then the unsupervised model can help to improve our estimates of $P(Y|X_C)$ and $P(X_E|Y)$, as demonstrated for the case of binary classification.
If, on the other hand, many possible $y$ can explain the observed $(x_C,x_E)$ equally well, then the unsupervised model appears to be less useful.



\subsubsection*{Acknowledgements}
The authors would like to thank Adrian Weller and Michele Tonutti for helpful feedback on the manuscript.

\bibliography{main}

\begin{thebibliography}{26}
\providecommand{\natexlab}[1]{#1}
\providecommand{\url}[1]{\texttt{#1}}
\expandafter\ifx\csname urlstyle\endcsname\relax
  \providecommand{\doi}[1]{doi: #1}\else
  \providecommand{\doi}{doi: \begingroup \urlstyle{rm}\Url}\fi

\bibitem[Blum and Mitchell(1998)]{blum1998combining}
A.~Blum and T.~Mitchell.
\newblock Combining labeled and unlabeled data with co-training.
\newblock In \emph{Proceedings of the eleventh annual conference on
  Computational learning theory}, pages 92--100. ACM, 1998.

\bibitem[Chapelle et~al.(2010)Chapelle, Sch{\"o}lkopf, and
  Zien]{Chapelle:2010:SL:1841234}
O.~Chapelle, B.~Sch{\"o}lkopf, and A.~Zien.
\newblock \emph{Semi-Supervised Learning}.
\newblock The MIT Press, 1st edition, 2010.

\bibitem[Fernando et~al.(2013)Fernando, Habrard, Sebban, and Tuytelaars]{SA}
B.~Fernando, A.~Habrard, M.~Sebban, and T.~Tuytelaars.
\newblock Unsupervised visual domain adaptation using subspace alignment.
\newblock In \emph{Proceedings of the IEEE international conference on computer
  vision}, pages 2960--2967, 2013.

\bibitem[Ganin et~al.(2016)Ganin, Ustinova, Ajakan, Germain, Larochelle,
  Laviolette, Marchand, and Lempitsky]{DANN}
Y.~Ganin, E.~Ustinova, H.~Ajakan, P.~Germain, H.~Larochelle, F.~Laviolette,
  M.~Marchand, and V.~Lempitsky.
\newblock Domain-adversarial training of neural networks.
\newblock \emph{Journal of Machine Learning Research}, 17\penalty0
  (59):\penalty0 1--35, 2016.

\bibitem[Gong et~al.(2012)Gong, Shi, Sha, and Grauman]{GFK}
B.~Gong, Y.~Shi, F.~Sha, and K.~Grauman.
\newblock Geodesic flow kernel for unsupervised domain adaptation.
\newblock In \emph{Computer Vision and Pattern Recognition (CVPR), 2012 IEEE
  Conference on}, pages 2066--2073. IEEE, 2012.

\bibitem[Janzing and Sch{\"o}lkopf(2010)]{janzing2010causal}
D.~Janzing and B.~Sch{\"o}lkopf.
\newblock Causal inference using the algorithmic markov condition.
\newblock \emph{IEEE Transactions on Information Theory}, 56\penalty0
  (10):\penalty0 5168--5194, 2010.

\bibitem[Janzing and Sch{\"o}lkopf(2015)]{janzing2015semi}
D.~Janzing and B.~Sch{\"o}lkopf.
\newblock Semi-supervised interpolation in an anticausal learning scenario.
\newblock \emph{Journal of Machine Learning Research}, 16:\penalty0 1923--1948,
  2015.

\bibitem[Koller and Friedman(2009)]{koller2009probabilistic}
D.~Koller and N.~Friedman.
\newblock \emph{Probabilistic graphical models: principles and techniques}.
\newblock MIT press, 2009.

\bibitem[Krogel and Scheffer(2004)]{krogel2004multi}
M.-A. Krogel and T.~Scheffer.
\newblock Multi-relational learning, text mining, and semi-supervised learning
  for functional genomics.
\newblock \emph{Machine Learning}, 57\penalty0 (1-2):\penalty0 61--81, 2004.

\bibitem[Pan and Yang(2010)]{pan2010survey}
S.~J. Pan and Q.~Yang.
\newblock A survey on transfer learning.
\newblock \emph{IEEE Transactions on knowledge and data engineering},
  22\penalty0 (10):\penalty0 1345--1359, 2010.

\bibitem[Pan et~al.(2011)Pan, Tsang, Kwok, and Yang]{TCA}
S.~J. Pan, I.~W. Tsang, J.~T. Kwok, and Q.~Yang.
\newblock Domain adaptation via transfer component analysis.
\newblock \emph{IEEE Transactions on Neural Networks}, 22\penalty0
  (2):\penalty0 199--210, 2011.

\bibitem[Patel et~al.(2015)Patel, Gopalan, Li, and Chellappa]{patel2015visual}
V.~M. Patel, R.~Gopalan, R.~Li, and R.~Chellappa.
\newblock Visual domain adaptation: A survey of recent advances.
\newblock \emph{IEEE signal processing magazine}, 32\penalty0 (3):\penalty0
  53--69, 2015.

\bibitem[Pearl(2009)]{pearl2009causality}
J.~Pearl.
\newblock \emph{Causality}.
\newblock Cambridge university press, 2009.

\bibitem[Peters et~al.(2016)Peters, B{\"u}hlmann, and
  Meinshausen]{peters2016invariant}
J.~Peters, P.~B{\"u}hlmann, and N.~Meinshausen.
\newblock Causal inference by using invariant prediction: identification and
  confidence intervals.
\newblock \emph{Journal of the Royal Statistical Society: Series B (Statistical
  Methodology)}, 78\penalty0 (5):\penalty0 947--1012, 2016.

\bibitem[Peters et~al.(2017)Peters, Janzing, and Sch{\"o}lkopf]{PetJanSch17}
J.~Peters, D.~Janzing, and B.~Sch{\"o}lkopf.
\newblock \emph{Elements of Causal Inference - Foundations and Learning
  Algorithms}.
\newblock Adaptive Computation and Machine Learning Series. The MIT Press,
  Cambridge, MA, USA, 2017.

\bibitem[Quionero-Candela et~al.(2009)Quionero-Candela, Sugiyama, Schwaighofer,
  and Lawrence]{quionero2009dataset}
J.~Quionero-Candela, M.~Sugiyama, A.~Schwaighofer, and N.~D. Lawrence.
\newblock \emph{Dataset shift in machine learning}.
\newblock The MIT Press, 2009.

\bibitem[Rojas-Carulla et~al.(2018)Rojas-Carulla, Sch{\"o}lkopf, Turner, and
  Peters]{rojas2018invariant}
M.~Rojas-Carulla, B.~Sch{\"o}lkopf, R.~Turner, and J.~Peters.
\newblock Invariant models for causal transfer learning.
\newblock \emph{Journal of Machine Learning Research}, 19\penalty0 (36), 2018.

\bibitem[Sachs et~al.(2005)Sachs, Perez, Pe'er, Lauffenburger, and
  Nolan]{sachs2005causal}
K.~Sachs, O.~Perez, D.~Pe'er, D.~A. Lauffenburger, and G.~P. Nolan.
\newblock Causal protein-signaling networks derived from multiparameter
  single-cell data.
\newblock \emph{Science}, 308\penalty0 (5721):\penalty0 523--529, 2005.

\bibitem[Sch{\"o}lkopf et~al.(2012)Sch{\"o}lkopf, Janzing, Peters, Sgouritsa,
  Zhang, and Mooij]{scholkopf2012causal}
B.~Sch{\"o}lkopf, D.~Janzing, J.~Peters, E.~Sgouritsa, K.~Zhang, and J.~Mooij.
\newblock On causal and anticausal learning.
\newblock In \emph{29th International Conference on Machine Learning (ICML
  2012)}, pages 1--8. International Machine Learning Society, 2012.

\bibitem[Shi and Sha(2012)]{ITL}
Y.~Shi and F.~Sha.
\newblock Information-theoretical learning of discriminative clusters for
  unsupervised domain adaptation.
\newblock In \emph{Proceedings of the 29th International Coference on
  International Conference on Machine Learning}, pages 1275--1282. Omnipress,
  2012.

\bibitem[Shimodaira(2000)]{shimodaira2000improving}
H.~Shimodaira.
\newblock Improving predictive inference under covariate shift by weighting the
  log-likelihood function.
\newblock \emph{Journal of statistical planning and inference}, 90\penalty0
  (2):\penalty0 227--244, 2000.

\bibitem[Storkey(2009)]{storkey2009training}
A.~Storkey.
\newblock When training and test sets are different: characterizing learning
  transfer.
\newblock \emph{Dataset shift in machine learning}, pages 3--28, 2009.

\bibitem[Sugiyama and Kawanabe(2012)]{sugiyama2012machine}
M.~Sugiyama and M.~Kawanabe.
\newblock \emph{Machine learning in non-stationary environments: Introduction
  to covariate shift adaptation}.
\newblock MIT press, 2012.

\bibitem[Sugiyama et~al.(2007)Sugiyama, Krauledat, and
  M{\"u}ller]{sugiyama2007covariate}
M.~Sugiyama, M.~Krauledat, and K.-R. M{\"u}ller.
\newblock Covariate shift adaptation by importance weighted cross validation.
\newblock \emph{Journal of Machine Learning Research}, 8\penalty0
  (May):\penalty0 985--1005, 2007.

\bibitem[Yan(2016)]{DAToolbox}
K.~Yan.
\newblock Domain adaptation toolbox.
\newblock \url{https://github.com/viggin/domain-adaptation-toolbox}, 2016.

\bibitem[Yan et~al.(2017)Yan, Kou, and Zhang]{MIDA}
K.~Yan, L.~Kou, and D.~Zhang.
\newblock Learning domain-invariant subspace using domain features and
  independence maximization.
\newblock \emph{IEEE transactions on cybernetics}, 2017.

\end{thebibliography}

\end{document}